\documentclass[letterpaper, 10 pt, conference]{ieeeconf}  

\IEEEoverridecommandlockouts                              

\usepackage{cite}
\usepackage{amsmath,amssymb,amsfonts}
\usepackage{algorithmic}
\usepackage{algorithm}
\usepackage{graphicx}
\usepackage{float}
\usepackage{textcomp}
\usepackage{xcolor}
\usepackage{multirow}
\usepackage{makecell}
\usepackage{colortbl}
\usepackage{xcolor}
\usepackage{bm}        

\title{A Fairness-Oriented Control Framework for Safety-Critical Multi-Robot Systems: Alternative Authority Control\\
\thanks{{*} indicates equal contribution. \ }

\thanks{ \noindent \textsuperscript{\textdagger} indicates corresponding author.\ }

\thanks{
L. Shi and Q.Liu are with University of Wisconsin-Madison,
        Madison, WI, US.{\tt\scriptsize \{
        lshi222, qliu426\} @wisc.edu}. \ 
        }
\thanks{
C.Zhou and X.Li are with Tencent Robotics X, Shenzhen, Guangdong, China. {\tt\scriptsize \{chowchzhou\} @tencent.com}. \ 
        }
\thanks{ Lei conducted this work during his internship at Tencent Robotics X.  \ }
}

\author{Lei Shi$^{*}$, Qichao Liu$^{*}$,  Cheng Zhou \noindent \textsuperscript{\textdagger}, Xiong Li}

\begin{document}

\maketitle
\thispagestyle{empty}
\pagestyle{empty}

\begin{abstract}
This paper proposes a fair control framework for multi-robot systems, which integrates the newly introduced Alternative Authority Control (AAC) and Flexible Control Barrier Function (F-CBF). Control authority refers to a single robot which can plan its trajectory while considering others as moving obstacles, meaning the other robots do not have authority to plan their own paths. The AAC method dynamically distributes the control authority, enabling fair and coordinated movement across the system. This approach significantly improves computational efficiency, scalability, and robustness in complex environments. The proposed F-CBF extends traditional CBFs by incorporating obstacle shape, velocity, and orientation. F-CBF enhances safety by accurate dynamic obstacle avoidance. The framework is validated through simulations in multi-robot scenarios, demonstrating its safety, robustness and computational efficiency.
\end{abstract}

\section{Introduction}

Multi-robot systems play a critical role in a variety of domains, including autonomous vehicular networks \cite{10571850}, automated warehousing, disaster response \cite{9220149}, and agriculture \cite{JU2022107336}. As these systems become more complex, ensuring both fair and efficient coordination among the robots is increasingly vital. In scenarios like multi-vehicle coordination or warehouse automation, uneven task distribution can overburden some robots while underutilizing others, reducing system efficiency and raising fairness concerns. In search and rescue, equitable control authority prevents monopolization, balances resource use, and improves search coverage, leading to more efficient and fair operations. 

Moreover, computational efficiency is essential for ensuring the reliability of multi-robot systems. In decentralized environments like warehouses, inefficient computation can result in delays, thereby decreasing productivity. Safety is another paramount consideration, particularly in dynamic environments where the presence of uncontrolled obstacles necessitates adaptive decision-making to prevent collisions and ensure smooth, safe operations.

This study focuses on the development of a control framework for efficient and fair authority distribution, improving dynamic obstacle avoidance in multi-robot systems. 

In summary, the key contributions of this paper are:

\begin{itemize}
    \item An Alternative Authority Control (AAC) framework is introduced, facilitating the dynamic distribution of control authority among robotic agents.
    
    \item A Model Predictive Control with Flexible Control Barrier Functions (MPC-FCBF) framework is proposed to enhance dynamic obstacle avoidance capabilities.
    
    \item A hierarchical control architecture is developed by integrating the AAC and MPC-FCBF frameworks, offering efficient dynamic authority allocation and improved obstacle avoidance for multi-agent robotic systems.
\end{itemize}

\section{Related Work}

Multi-robot control is categorized into centralized and decentralized paradigms \cite{1068004}. While centralized control excel in global optimization, they face challenges in computational efficiency and scalability as complexity rises. Decentralized frameworks are scalable and robust against robot failures or communication issues \cite{6548083}. They adapt well to complex environments, and are widely used in systems like UAVs, UGVs, and unmanned marine vehicles. Decentralized strategies like virtual structures \cite{lewis1997high} and leader-follower models \cite{8957499} are popular for their robustness and scalability. The leader-follower control (LFC) approach is widely used for its simplicity and effectiveness in multi-robot coordination. Additionally, decentralized control can handle complex tasks using Lyapunov-like barrier functions \cite{7122266}. While RL methods can learn fair policies \cite{zimmer2021learning}, their transferability is limited. A fairness-focused control framework, combined with other control algorithms, could greatly improve multi-robot control in various environments.

Control Barrier Functions (CBFs) are a key tool for robot safety, transforming safety control into a constrained optimal control problem for real-time obstacle avoidance and energy efficiency \cite{ames2016control,ames2019control}. Recent advancements have introduced specialized CBFs, such as Visual CBFs \cite{10161482,10160805} and Risk-Aware CBFs \cite{10341919,10161379}, enhancing performance in visual feedback control and risk perception. Integration with Reinforcement Learning has also shown promise in improving safety for autonomous driving \cite{10161418,10160991}.

In multi-obstacle environments, researchers have developed multi-constrained CBFs \cite{10160974} and combined Dynamic CBFs with Model Predictive Control \cite{10160857}. However, these methods, including D-CBFs that rely solely on relative state information, do not always guarantee absolute safety, particularly when dealing with high-speed obstacles. To address these limitations, we propose a novel Alternative Authority Control (AAC) framework to enhance the fairness, robustness, and computational efficiency of multi-robot systems, along with a Flexible Control Barrier Function (F-CBF) to improve safety and adaptability in decentralized control systems.

\section{Upper Framework: Alternative Authority Control}



\subsection{Overview of AAC Framework}

\begin{figure}[htbp]
    \centering
    \includegraphics[width=3.3 in]{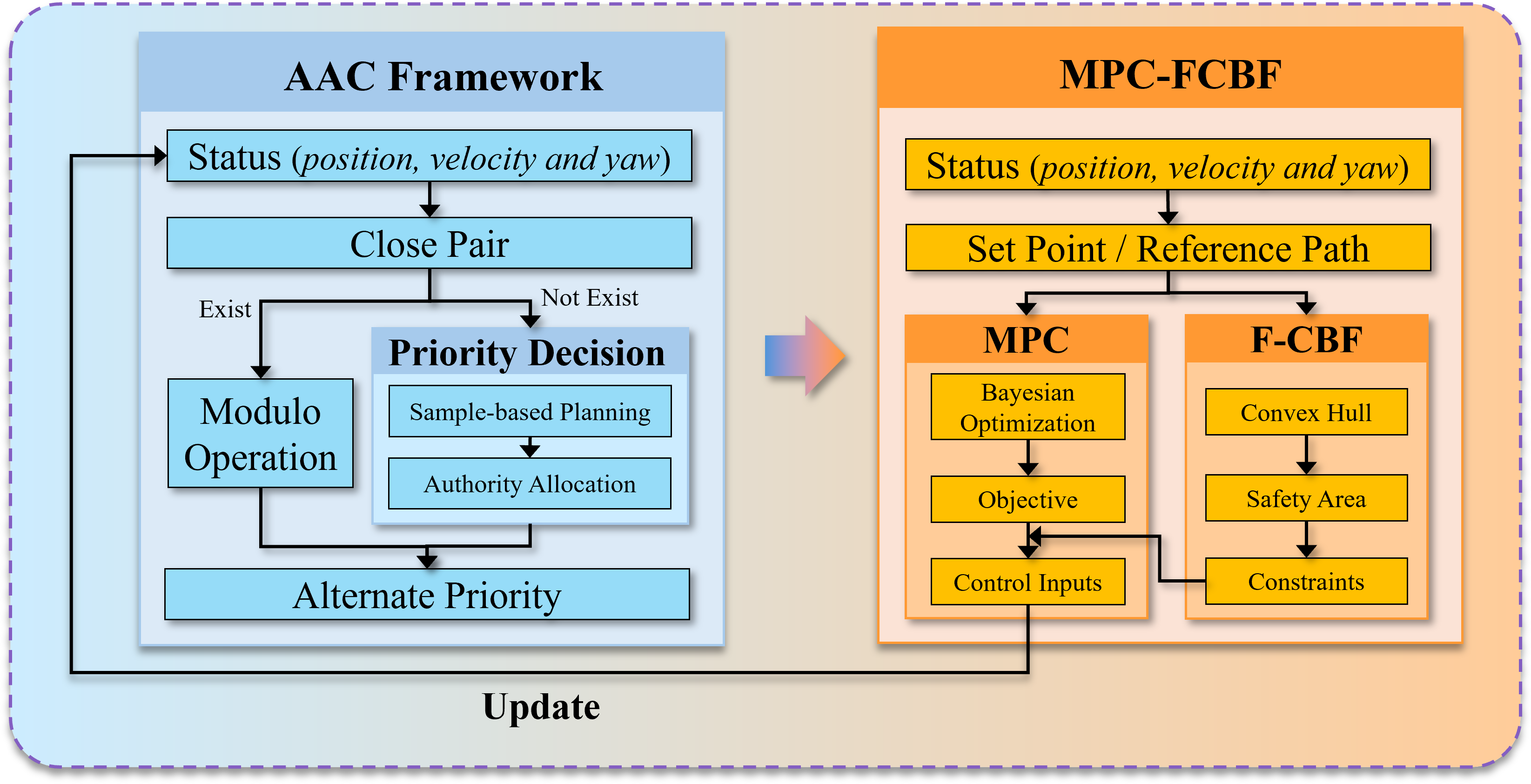}
    \caption{Overview of system framework}
    \label{fig:wide_figure}
\end{figure}

AAC is an innovative approach to multi-robot coordination that addresses the limitations of traditional centralized control systems. This method enables more flexible, efficient, and robust multi-robot system control through the dynamic allocation of control authority—i.e., the responsibility for path planning—among the robots in the system.

The core idea of AAC is that at each time step, control authority is assigned to a single robot, termed the ``authority robot". This robot is responsible for planning its own path while considering the other robots as dynamic obstacles. The authority robot utilizes the MPC-FCBF method to generate a safe path that avoids collisions. Meanwhile, the non-authority robots are treated as obstacles and contribute to the path planning by generating F-CBF constraints. This approach ensures that each robot is accounted for during the planning process, leading to more efficient and coordinated motion. The AAC pseudocode is presented in Algorithm 1:

\begin{algorithm}[htbp]
    \footnotesize
    \caption{Alternative Authority Control}\label{alg:alternative_authority_control}
    \begin{algorithmic}
    \STATE \textbf{Given:} $robots$ - list of robots
    \STATE \textbf{Initialize:} $robots.position \gets initial\_positions$, $robots.path \gets \emptyset$
    \STATE
    \STATE \textbf{for} $iteration = 1$ \textbf{to} $num\_iterations$ \textbf{do}
    \STATE \hspace{0.4cm} $close\_pairs \gets$ Identify pairs of robots that are close to each other
    \STATE \hspace{0.4cm} \textbf{if} $close\_pairs \neq \emptyset$ \textbf{then}
    \STATE \hspace{0.8cm} $authority\_robot \gets$ \textsc{Proximity-Decision}$(close\_pairs)$
    \STATE \hspace{0.4cm} \textbf{else}
    \STATE \hspace{0.8cm} $authority\_robot \gets robots[iteration \bmod |robots|]$
    \STATE \hspace{0.4cm} \textbf{end if}
    \STATE \hspace{0.4cm} $obstacles \gets$ \textsc{ConvexHull}$(Non\text{-}authority\ robots)$
    \STATE \hspace{0.4cm} $robots.move \gets$ \textsc{MPC-FCBF}$(authority\_robot, obstacles)$
    \STATE \textbf{end for}
    \STATE 
    \STATE \textbf{Function} \textsc{Proximity-Decision}$(robots)$:
    \STATE \hspace{0.4cm} $best\_overall\_path \gets \emptyset$, $best\_robot \gets \texttt{null}$
    \STATE \hspace{0.4cm} \textbf{for each} $robot$ \textbf{in} $robots$ \textbf{do}
    \STATE \hspace{0.8cm} Initialize tree $T$ with $robot$'s current state as root; $best\_path \gets \emptyset$
    \STATE \hspace{0.8cm} \textbf{for} $i = 1$ \textbf{to} $num\_proximity \_iterations$ \textbf{do}
    \STATE \hspace{1.2cm} $q_{rand} \gets$ GenerateRandomState()
    \STATE \hspace{1.2cm} $q_{near} \gets$ FindNearestNode$(T, q_{rand})$
    \STATE \hspace{1.2cm} $q_{new} \gets$ CreateNewNode$(q_{near}, q_{rand})$
    \STATE \hspace{1.2cm} $score \gets$ CalculateScore$(q_{new})$
    \STATE \hspace{1.2cm} \textbf{if} $score > $ Score$(best\_path)$ \textbf{then}
    \STATE \hspace{1.6cm} $best\_path \gets$ GeneratePath$(T.root, q_{new})$
    \STATE \hspace{1.2cm} \textbf{end if}
    \STATE \hspace{0.8cm} \textbf{end for}
    \STATE \hspace{0.8cm} \textbf{if} Progress$(best\_path) >$ Progress$(best\_overall\_path)$ \textbf{then}
    \STATE \hspace{1.2cm} $best\_overall\_path \gets best\_path$
    \STATE \hspace{1.2cm} $best\_robot \gets robot$
    \STATE \hspace{0.8cm} \textbf{end if}
    \STATE \hspace{0.4cm} \textbf{end for}
    \STATE \hspace{0.4cm} \textbf{return} $best\_robot$
    \STATE
    \STATE \textbf{Function} \textsc{ConvexHull}$(obstacles)$:
    \STATE \hspace{0.4cm} Initialize $hull$ as an empty set
    \STATE \hspace{0.4cm} \textbf{for each} $obstacle \in obstacles$ \textbf{do}
    \STATE \hspace{0.8cm} Add current and next positions of $obstacle$ to $hull$
    \STATE \hspace{0.4cm} \textbf{end for}
    \STATE \hspace{0.4cm} Compute $convex\_hull$ from $hull$
    \STATE \hspace{0.4cm} \textbf{return} $convex\_hull$
    \end{algorithmic}
\end{algorithm}

\subsection{Advantages} 
\begin{itemize}
    \item \textbf{Computational Efficiency and Scalability:} The AAC framework assigns control authority to one robot at a time, reducing computational complexity by focusing resources on a single decision-maker. This improves computational efficiency and allows the system to scale, even with many robots.
    
    \item \textbf{Robustness:} The dynamic allocation of control authority ensures smooth traffic flow, preventing conflicts and deadlocks. Additionally, it maintains greater inter-robot distances, reducing the risk of collisions.
    
    \item \textbf{Fairness:} The AAC framework ensures fairness by rotating control authority among all robots, giving each an equal opportunity to plan and execute its path. This prevents any single robot from monopolizing decision-making or resources, ensuring that all robots have a more balanced distribution. 
\end{itemize}

\subsection{Proximity Decision and Predictive Collision Avoidance}
\subsubsection{\textbf{Proximity Decision Function}}
Simply rotating authority among robots using a remainder approach can lead to inefficiencies, particularly when robots are in close proximity. In such cases, nearby robots would repeatedly exchange authority, resulting in a deadlock where neither effectively yields to the other. This potential stalemate necessitates a more sophisticated approach to authority allocation in close pairs.

The Proximity Decision Function determines the authority allocation of close pairs using a sampling-based method \cite{lavalle2001rapidly} to evaluate each robot's potential paths. For each robot \(R_i\), the function initializes a tree structure \(\mathcal{T}_i\) with the robot's current state \(s_i^0\). The tree explores potential future states through random sampling. In each iteration, it generates a random state \(s_i^{\text{rand}}\), finds the nearest existing node \(s_i^{\text{near}}\) in the tree, and creates a new node \(s_i^{\text{new}}\). A score \(f(s_i^{\text{new}})\) is calculated for this new node, which balances progress towards the goal and collision avoidance:
\begin{equation}
f(s_i^{\text{new}}) = \alpha \cdot d(s_i^{\text{new}}, g_i) - \beta \cdot \sum_{j \neq i} C(s_i^{\text{new}}, s_j)
\end{equation}
where \(d(s_i^{\text{new}}, g_i)\) is the distance from the new state to the goal \(g_i\), \(C(s_i^{\text{new}}, s_j)\) is a collision cost with another robot \(R_j\), and \(\alpha\) and \(\beta\) are weighting factors.

The function evaluates each new state \(s_i^{\text{new}}\), updating the robot's best path \(\pi_i\) if with a higher score \(f(s_i^{\text{new}})\). The best paths of all robots are then compared, with the authority allocated to the robot \(R_{i^*}\) whose best path \(\pi_{i^*}\) shows the greatest progress. This allocation is formally expressed as: $i^* = \arg \max_{i} f(\pi_i)$. This method assigns authority based on exploring each robot's potential paths, enhancing robustness and avoiding deadlocks.

\subsubsection{\textbf{Predictive Collision Avoidance}}

The Convex Hull function \cite{10.1145/235815.235821} is used to create a simplified representation of each non-authority robots. Let \(\mathbf{p}_i(t)\) denote the current position of the \(i\)-th non-authority robot at time \(t\), and \(\mathbf{p}_i(t+1)\) denote its predicted position at time \(t+1\). The convex hull \(\text{CH}(\cdot)\) is computed over both current and predicted next positions of each non-authority robot. Mathematically, for the \(i\)-th robot, this can be expressed as:
\begin{equation}
 \text{CH}_i = \text{ConvexHull}(\{\mathbf{p}_i(t), \mathbf{p}_i(t+1)\})   
\end{equation}

This convex hull \(\text{CH}_i\) represents the area that the \(i\)-th robot occupy from time \(t\) to \(t+1\).

The authority robot, with position \(\mathbf{q}(t)\) at time \(t\), must avoid these dynamic obstacles. To ensure collision avoidance, the trajectory of the authority robot is planned such that:
\begin{equation}
 \mathbf{q}(t+1) \notin \bigcup_{i} \text{CH}_i   
\end{equation}


By including predicted next positions \(\mathbf{p}_i(t+1)\) in the convex hull, the system avoids both current and future obstacles, allowing both authority and non-authority robots to safely move in a single iteration. In contrast, considering only current obstacle positions could lead to collisions between the authority robot and the non-authority robots in the next step. Moving all robots together advances the system further towards the goal, reducing the total iterations needed and improving computational efficiency while ensuring robust collision avoidance.

\section{Lower Controller: MPC-FCBF}

\subsection{Flexible Control Barrier Function}

Control Barrier Functions (CBFs) ensure safety by avoiding obstacles in autonomous systems. They work well with stationary objects but struggle with moving ones, limiting their effectiveness in dynamic environments. We propose Flexible Control Barrier Functions (F-CBFs) that improve dynamic obstacle avoidance by considering the state, velocity, and shape of obstacles. Define robot position, obstacles position, obstacles shape as \(\mathbf{x} \in \mathbb{R}^n \),  \(\mathbf{x}_{ob} \in \mathbb{R}^m \) and \(\mathbf{\eta}_{ob} \in \mathbb{R}^p\), and define environmental variable \(\mathbf{X}= [x, x_{ob}, \eta_{ob}] \in \mathbb{R}^n \times \mathbb{R}^m \times \mathbb{R}^p\).

To ensure the safety of system operations, we define a safe set \( C \) that includes all states \( X \) satisfying the condition \( h(X) \geq 0 \), where \( h \) is a continuously differentiable function and \( \mathcal{X} \) is a subset of the state space:

\begin{equation}
    C = \{X \in \mathcal{X} : h(X) \geq 0\}
    \label{eq:1}
\end{equation}




\begin{figure}[htbp]
\centerline{\includegraphics[width=3.2 in]{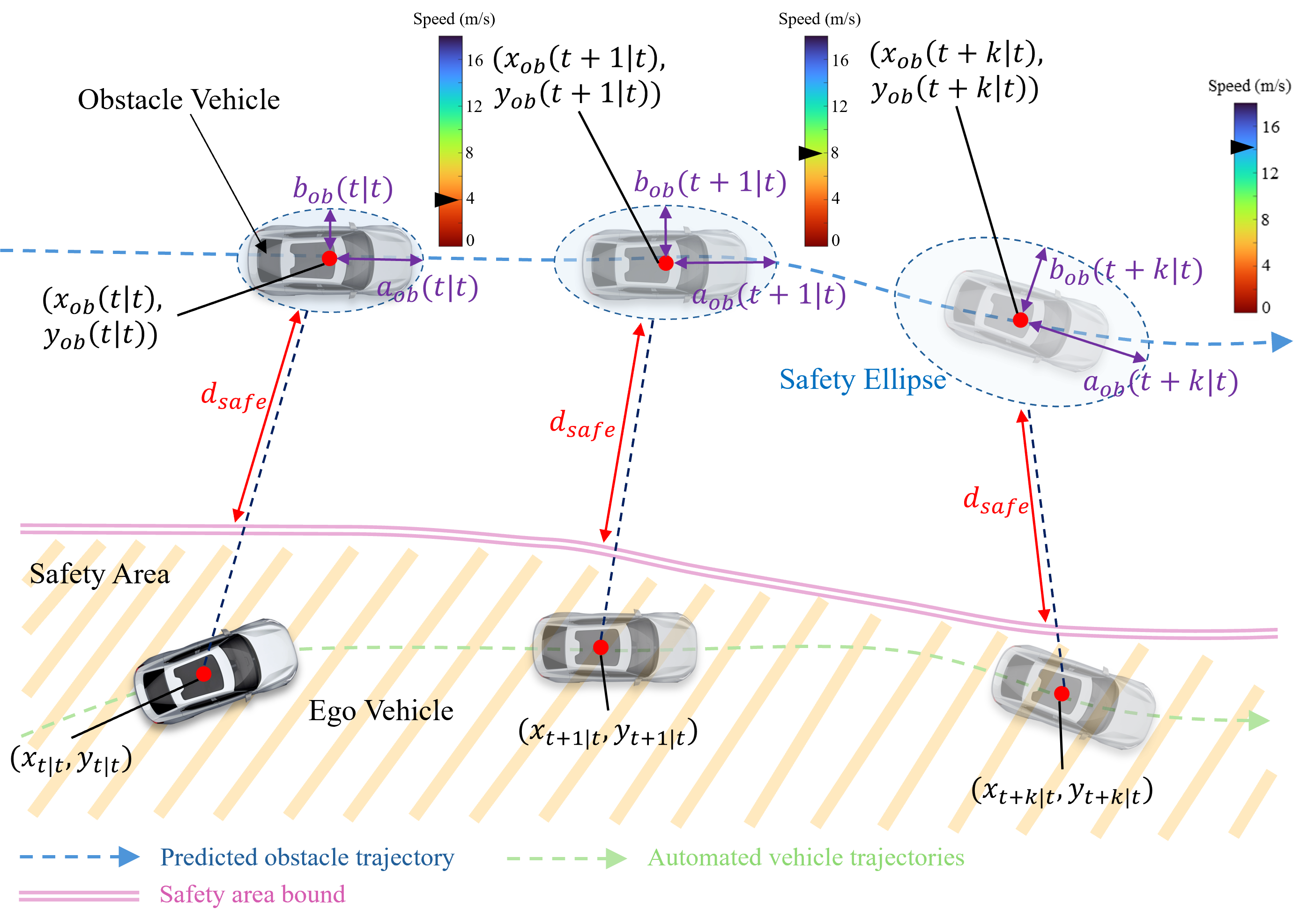}}
\caption{F-CBF Illustration: By continuously monitoring the speed and position of dynamic obstacles, the size and orientation of the safety ellipse are dynamically adjusted to modify the boundaries of the safety region.}
\label{fig}
\end{figure}

We define the F-CBF as an extension of the traditional CBF. Let \(\eta_{ob} = [\eta_1, \eta_2, \eta_3]^T\) represent the obstacle's shape, where:

\begin{itemize}
    \item We represent both the obstacle and the robot using ellipses to capture their shape and movement. In this representation, \(\eta_1\) and \(\eta_2\) describe the semi-major and semi-minor axes of the elliptical obstacle, respectively, and are functions of the obstacle's dimensions and speed.
    \item \(\eta_3\) represents the orientation of the ellipse, determined by the obstacle's velocity direction.
\end{itemize}

For a F-CBF \(h\), we require that for all \(X \in \partial C\), \(\frac{\partial h}{\partial X} \neq 0\), and there exists an extended class \(\mathcal{K}_\infty\) function \(\gamma\) such that:
\begin{equation}
    \exists u \; \text{s.t.} \; h'(X, u, \eta_{ob}) \geq -\gamma(h(X)), \quad \gamma \in \mathcal{K}_\infty
    \label{eq:2}
\end{equation}

The time derivative of \(h\) is given by:
\begin{equation}
    h'(X, u, \eta_{ob}) = \frac{\partial h}{\partial X} \dot{X} + \frac{\partial h}{\partial \eta_{ob}} \dot{\eta}_{ob}
    \label{eq:3}
\end{equation}
where \(\dot{\eta}_{ob}\) represents the rate of change in the obstacle's shape and orientation, a critical aspect of F-CBF.

For input-affine systems \(\dot{x} = f(x) + g(x)u\), the F-CBF condition from Eq. \ref{eq:2} can be rewritten as:
\begin{equation}
\sup_{u \in \mathcal{U}} \left[ L_f h + L_g h u + \underbrace{\frac{\partial h}{\partial x_{ob}} \frac{\partial x_{ob}}{\partial t}}_{\epsilon} + \underbrace{\frac{\partial h}{\partial \eta_{ob}} \frac{\partial \eta_{ob}}{\partial t}}_{\rho} \right] + \gamma h \geq 0
\label{eq:3}
\end{equation}

Here, \(\epsilon\) represents the impact of the obstacle's state, and \(\rho\), which directly involves \(\eta\), captures the influence of the obstacle's shape and its changes on the control input.

The control set \(K_{fcbf}\) for F-CBF is defined as:
\begin{equation}
K_{f_{cbf}} = \left\{ u : L_f h + L_g h u + \epsilon + \rho + \gamma h \geq 0 \right\}
\label{eq:4}
\end{equation}

Comparing F-CBF with traditional CBF:
\begin{itemize}
    \item If \(\epsilon + \rho < 0\), \(K_{f_{cbf}}\) is a subset of the CBF control set, implying stricter constraints due to \(\eta\).
    \item If \(\epsilon + \rho > 0\), the CBF control set is a subset of \(K_{f_{cbf}}\), showing that F-CBF imposes stricter constraints because of \(\eta\).
    \item If \(\epsilon + \rho = 0\), both F-CBF and CBF impose the same constraints.
\end{itemize}

For discrete-time systems, Eq. \ref{eq:2} is rewritten as:
\begin{equation}
    \Delta h(X_k, u_k) \geq -\gamma h(X_k), \quad 0 < \gamma \leq 1
    \label{eq:5}
\end{equation}
where \(\Delta h(X_k, u_k) := h(X_{k+1}) - h(X_k)\).

The semi-major (\(a_{ob}\)) and semi-minor (\(b_{ob}\)) axes of the obstacle's safety ellipse are defined as:
\begin{align}
a_{ob} &= L_{ob}/2 + \mu_1  \cdot \sigma(|\vec{v}_{ob,\parallel}| - v_0) \cdot |\vec{v}_{ob,\parallel}| \label{eq:6}\\
b_{ob} &= W_{ob}/2 + \mu_2  \cdot \sigma(|\vec{v}_{ob,\perp}| - v_0) \cdot |\vec{v}_{ob,\perp}| \label{eq:7}
\end{align}
Here, \(L_{ob}\) and \(W_{ob}\) represent the obstacle's length and width, respectively, and \(\sigma(x)\) is the sigmoid function: 

\begin{equation}
    \sigma(x) = \frac{1}{1 + e^{-kx}}
\end{equation}
\(v_0\) is the velocity threshold, and \(k\) is the steepness parameter of the sigmoid function. \(\mu_1\) and \(\mu_2\) adjust the correlation between the ellipse's shape and the obstacle's speed. \( \vec{v}_{ob,\parallel} \) and \( \vec{v}_{ob,\perp} \) represent the velocity components along the major and minor axes, respectively. The standard equation of the safety ellipse in the local coordinate system is:
\begin{equation}
    \frac{x'^2}{a_{ob}^2} + \frac{y'^2}{b_{ob}^2} = 1
\label{eq:8}
\end{equation}

To represent points on the ellipse, we use the parameter \(\sigma\) ranging from \(0\) to \(2\pi\), with the following parametric equations:
\begin{align}
x'(\sigma) &= a_{ob} \cos(\sigma) \label{eq:9}\\
y'(\sigma) &= b_{ob}\sin(\sigma)\label{eq:10}
\end{align}

Obstacles move in a two-dimensional plane with changing speed and direction. Since Eq. \ref{eq:6} and \ref{eq:7} consider only speed variation, we introduce the rotation matrix \( R(\phi(t))\) to align the ellipse with the obstacle's rotation angle $\phi(t)$. This ensures the semi-major axis always points in the direction of movement. The transformation is shown in Eq. \ref{eq:11}.
\begin{equation}
\begin{bmatrix}
x_e(\sigma) \\
y_e(\sigma)
\end{bmatrix}
= \underbrace{\begin{bmatrix}
\cos(\phi) & -\sin(\phi) \\
\sin(\phi) & \cos(\phi)
\end{bmatrix}}_{R(\phi)}
\begin{bmatrix}
x'(\sigma) \\
y'(\sigma)
\end{bmatrix}
+
\begin{bmatrix}
x_{ob} \\
y_{ob}
\end{bmatrix}
\label{eq:11}
\end{equation}

Substituting the parametric equations into Eq. \ref{eq:11}:
\begin{align}
x_e(\sigma) &= x_{ob}+ a_{ob} \cos(\sigma) \cos(\phi) - b_{ob} \sin(\sigma) \sin(\phi) \label{eq:13}\\
y_e(\sigma) &= y_{ob} + a_{ob} \cos(\sigma) \sin(\phi) + b_{ob} \sin(\sigma) \cos(\phi) \label{eq:14}
\end{align}

Define \( p(k) = [x(k), y(k)] \) to represent the position of the autonomous robot. Therefore, the shortest distance between the autonomous robot and the dynamic safety ellipse of the dynamic obstacle can be determined by solving the following optimization problem.
\begin{equation}
    \sigma_{\text{min}} = \arg \min_{\sigma \in [0, 2\pi]} d(\sigma)
\end{equation}
where \(d(\sigma)\) is the Euclidean distance from the ego robot to any point on the dynamic safety ellipse, calculated as follows:
\begin{equation}
  d(\sigma) = \sqrt{(x_e(\sigma) - x(k))^2 + (y_e(\sigma) - y(k))^2} 
\end{equation}

Thus, F-CBF is formulated in a quadratic form as follows:
\begin{equation}
\begin{aligned}
    h(X_k) &= d(\sigma_{min}) - d_{\text{safe}}, &\quad k = 0, 1, \ldots, N-1
\end{aligned}
\end{equation}

\subsection{Modelling of MPC-FCBF}

To formulate MPC, it is essential to first describe the robot's dynamic model. This dynamic model is articulated through a discrete-time equation, \( \mathbf{x}_{k+1} = f(\mathbf{x}_k, \mathbf{u}_k) \). We use the bicycle model to describe the vehicle's dynamics. This model simplifies the vehicle's motion into a two-dimensional framework, retaining essential characteristics while reducing computational complexity. The discrete form of the bicycle model is given by:
\begin{equation}
\begin{cases}
x_i(k+1) = x_i(k) + v_i(k) \cos \varphi_i(k) \Delta t \\
y_i(k+1) = y_i(k) + v_i(k) \sin \varphi_i(k) \Delta t \\
\varphi_i(k+1) = \varphi_i(k) + \frac{v_i(k) \tan \delta_i(k)}{L} \Delta t \\
v_i(k+1) = v_i(k) + a_i(k) \Delta t
\end{cases}
\end{equation}

Here, \((x_i, y_i)\) are the coordinates of the midpoint of vehicle \(i\)'s rear axle; \(\varphi_i\) is the yaw angle; \(v_i\) is the longitudinal velocity; \(L\) is the wheelbase; \(\delta_i\) is the steering angle; and \(a_i\) is the longitudinal acceleration.


During the control process, it is necessary to ensure that the robot meets the terminal state requirements and follows the planned path. We define \( p(\mathbf{x}_{t+N|t}) \) and \( q(\mathbf{x}_k, \mathbf{u}_k)\) as the terminal cost and running cost of MPC problem, respectively. The specific forms are given in Eq.s \ref{eq:23} to \ref{eq:24}:
\begin{equation}
p(\mathbf{x}_{t+N|t}) = \mathbf{x}_f^T \mathbf{P} \mathbf{x}_f
\label{eq:23}
\end{equation}
\begin{equation}
q(\mathbf{x}_k, \mathbf{u}_k) = (\mathbf{x}_k - \mathbf{T}_k)^T \mathbf{Q} (\mathbf{X}_k - \mathbf{T}_k) + \mathbf{u}_k^T \mathbf{R} \mathbf{u}_k
\label{eq:24}
\end{equation}

The MPC control problem is formulated as:
\begin{subequations}
\small
\begin{align}
    J_t^*(\mathbf{x}_t) = & \min_{\mathbf{u}_t:t+N-1|t} p(\mathbf{x}_{t+N|t}) + \sum_{k=0}^{N-1} q(\mathbf{x}_k, \mathbf{u}_k) \label{eq:25a}\\
    \text{s.t.} \quad & \mathbf{x}_{k+1} = f(\mathbf{x}_k, \mathbf{u}_k), \quad k = 0, 1, \ldots, N-1 \label{eq:25b}\\
    & \mathbf{x}_k \in \bm{\mathcal{X}}, \mathbf{u}_k \in \bm{\mathcal{U}}, \quad k = 0, 1, \ldots, N-1 \\
    & \mathbf{x}_{t|t} = \mathbf{x}_t \label{eq:25d}\\
    & \mathbf{x}_{t+N|t} \in \bm{\mathcal{X}}_f \\
    & \Delta h(\mathbf{X}_k, \mathbf{u}_k) \geq -\gamma h(\mathbf{X}_k), \quad k = 0, 1, \ldots, N-1
\end{align}
\end{subequations}

In this context, Eq. \ref{eq:25b} describes the system dynamics, while Eq. \ref{eq:25d} represents the initial condition at step \( t \in \mathbb{Z}^+ \). The sets \( \mathcal{X} \), \( \mathcal{U} \), and \( \mathcal{X}_f \) denote the reachable states, inputs, and terminal states, respectively. To track the reference path \( x_{t+N|t}^d \) from the global planner, the terminal cost \( p(x_{t+N|t}) \) in Eq. \ref{eq:25a} is defined as \( \|x_{t+N|t} - x_{t+N|t}^d\|_P \), and the stage cost \( q(x_k, u_k) \) is defined as \( \|x_k - x_k^d\|_Q + \|u_k\|_R + \|u_k - u_{k-1}\|_S \), where \( P \), \( Q \), \( R \), and \( S \) are weight matrices.

\subsection{Parameter Tuning via Bayesian Optimization}

Bayesian optimization is used to tune the MPC parameters due to the complexity of finding the optimal settings \cite{lu2020mpc}. The optimization problem is defined as:
\begin{equation}
    \theta^* = \arg \max_{\theta \in \Theta} f(\theta)
\end{equation}
where \( f(\theta) \) is the objective function, and \( \theta \) includes the MPC parameters such as the horizon and weights. In each iteration, the next point \( \theta_{n+1} \) is chosen by maximizing the acquisition function \( a(\theta) \), specifically the Expected Value:
\begin{equation}
    \theta_{n+1} = \arg\max_{\theta} \, a(\theta) = \arg\max_{\theta} \, \mathbb{E} \left[\max(f(\theta) - f(\theta^+), 0)\right]
\end{equation}
where \( f(\theta^+) \) is the best value observed so far. This method effectively find the optimal MPC parameters.

\section{Experiments}
\subsection{AAC Framework Validation and Analysis}

\begin{figure}[H]
    \centering
    \includegraphics[width=3.2 in]{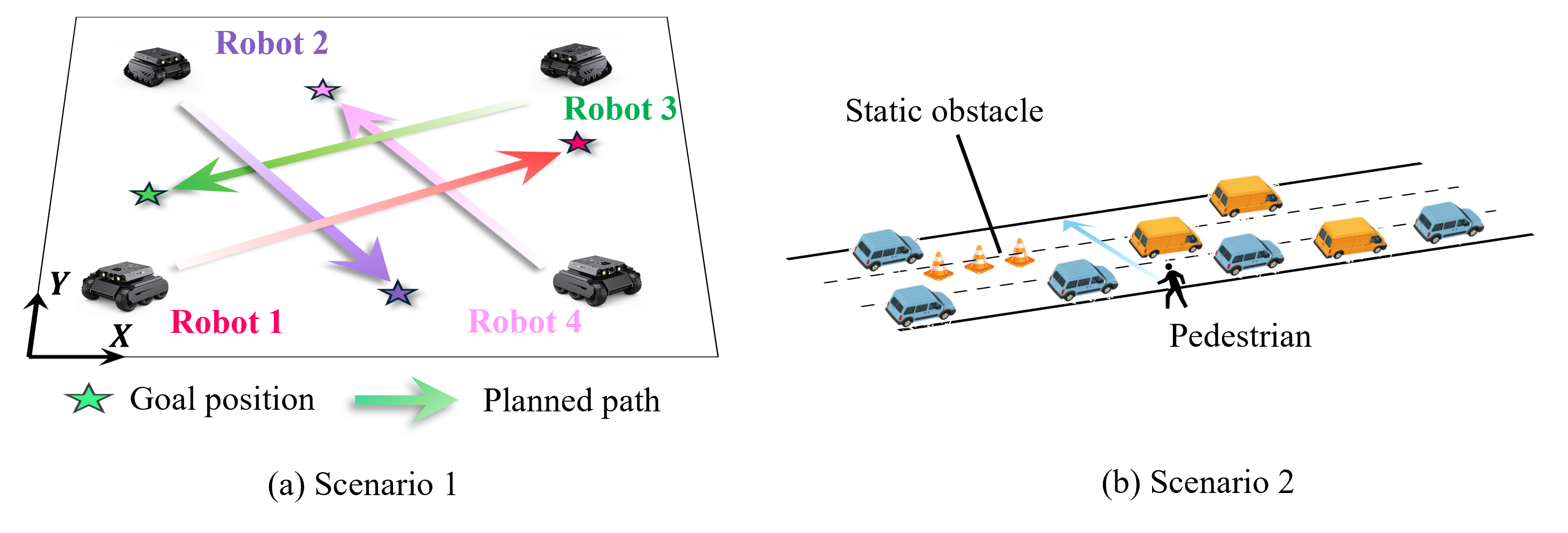}
    \caption{Test scenarios for swarm robots and autonomous driving}
    \label{fig:scenario}
\end{figure}

In this section, we validate the AAC framework and the MPC-FCBF control strategy for multi-robot control tasks in the two scenarios as shown in figure/Fig.\ref{fig:scenario}:

\begin{itemize}
    \item \textbf{Scenario 1:} In the search and rescue task, the MPC controller was designed to minimize the Euclidean distance between the rescue robot and the location of the trapped individuals, ensuring timely and accurate navigation through complex environments.
    \item \textbf{Scenario 2:} In a real-world scenario with static obstacles (traffic cones) and dynamic obstacles (uncontrolled pedestrians), 15 vehicles must complete coordinated lane changes. Each vehicle follows a pre-planned trajectory that doesn't consider collisions with other vehicles or obstacles.
\end{itemize}

This study selects Leader-Follower Control (LFC) as the benchmark for its fast computation and high stability in multi-robot control. In our LFC, robots with smaller index numbers have higher priority and take on the leader role.

\begin{figure}[htbp]
    \centering
    \includegraphics[width=2.5 in]{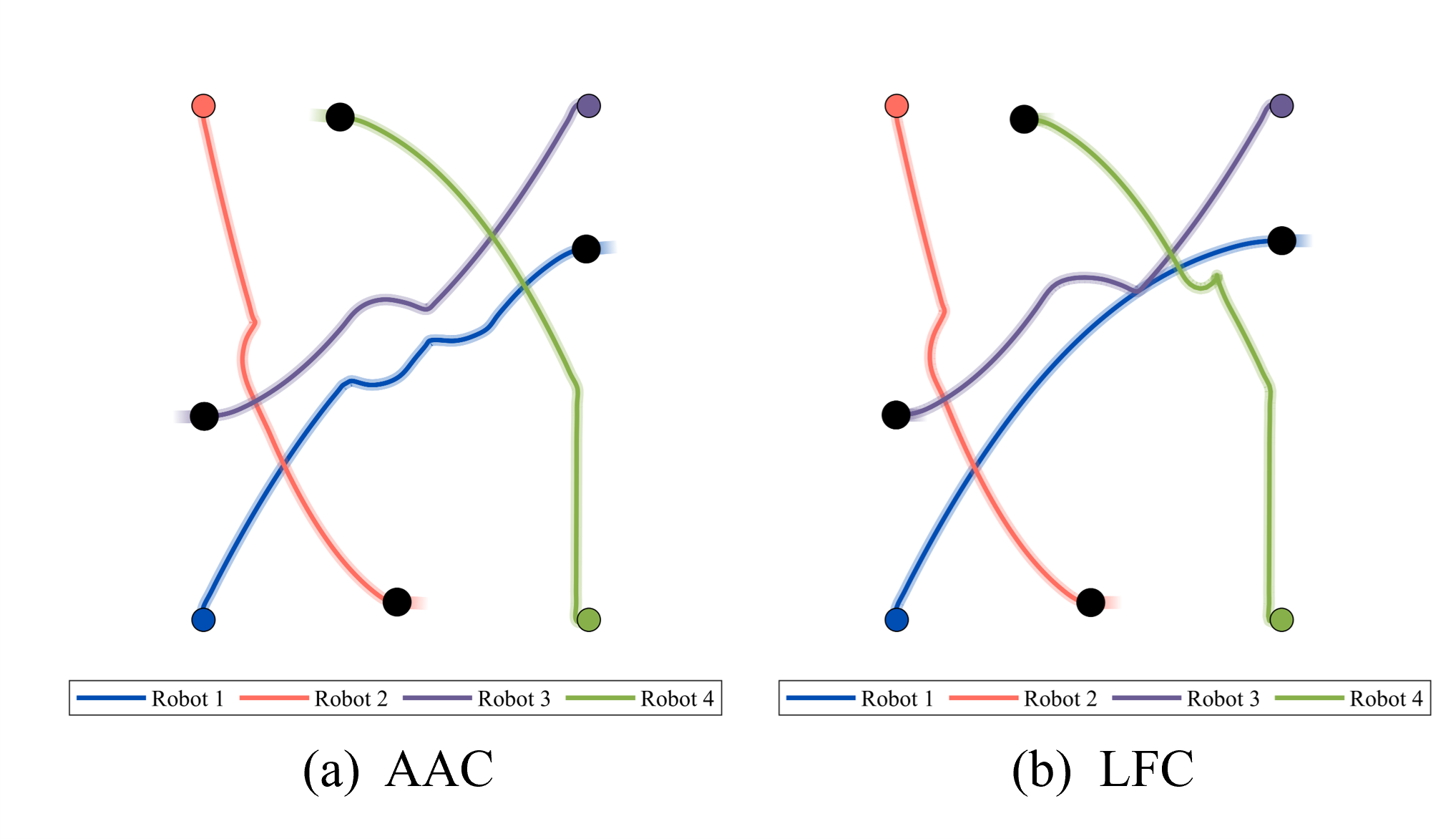}
    \caption{Comparison of target-reaching trajectories for wheeled robots under two frameworks}
    \label{fig:UGV results}
\end{figure}

Table \ref{tab1} and figure/Fig. \ref{fig:UGV results} compare the performance of the AAC and LFC methods in Scenario 1. As shown in figure/Fig. \ref{fig:UGV results}(b), the LFC method results in abrupt changes and unsmooth trajectories because robots 3 and 4 need to wait for the higher-priority robot 1 to pass, and robot 4 also has to wait for robot 3 before it can proceed. In contrast, figure/Fig. \ref{fig:UGV results}(a) demonstrates that the AAC framework effectively avoids such issues. The smooth trajectories not only prevent stalling for Robot 4, which would otherwise have to yield to other agents, but also highlight the framework's inherent fairness. The fair distribution of control authority ensures no single robot dominates the shared space, enhancing overall system efficiency. This efficiency is shown in the consistently shorter Task Duration of AAC, demonstrating how fairness improves the system's overall performance. The statistical results further support this conclusion: the AAC method reduces task duration time by approximately 11\% and computation time by about 26\%. Moreover, the AAC framework shows an advantage in robustness by maintaining a larger minimum distance, which poses a lower collision risk.

\begin{figure*}[htbp]
    \centering
    \includegraphics[width=\textwidth]{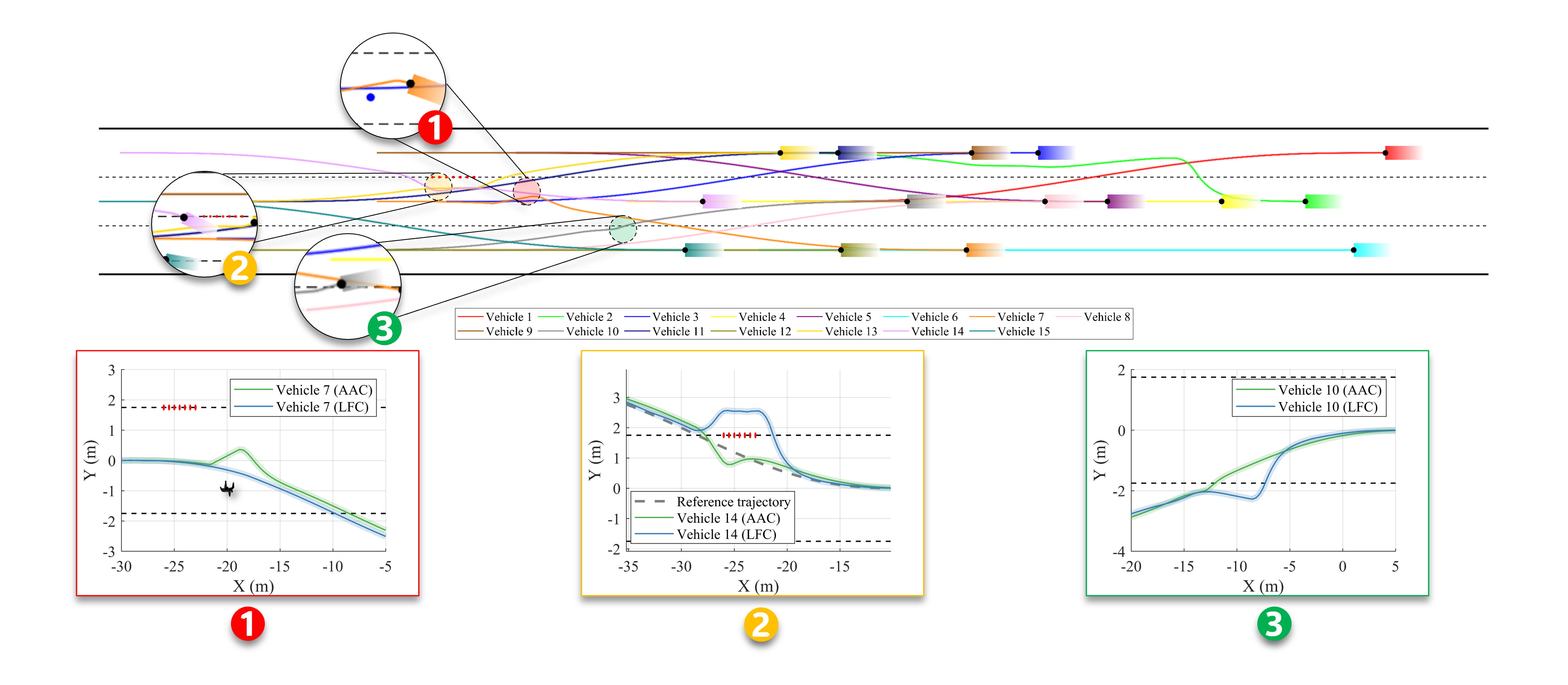}
    \caption{A comparison of trajectories with AAC and LFC}
    \label{fig:AV results}
\end{figure*}

\begin{table}[htbp]
\caption{Statistical Results for Scenario 1}
\centering
\footnotesize
\setlength{\tabcolsep}{3pt}
\begin{tabular}{|c|c|p{1.5cm}|p{1.5cm}|p{1.5cm}|}
\hline
\textbf{Controller} & \textbf{N} & \textbf{Task Duration [s]} & \textbf{Computation Time [ms]} & \textbf{Minimum Dist. [m]} \\ \hline
AAC-MPC-FCBF & \multirow{2}{*}{5}  & 18.3 & 58.7 & 1.19 \\ \cline{1-1}\cline{3-5} 
LFC-MPC-FCBF &  & 22.6 & 79.6 & 0.47 \\ \hline
AAC-MPC-FCBF & \multirow{2}{*}{10} & 18.1 & 68.4 & 1.13 \\ \cline{1-1}\cline{3-5} 
LFC-MPC-FCBF &  & 21.2 & 125.6 & 0.48 \\ \hline
AAC-MPC-FCBF & \multirow{2}{*}{15} & 17.6 & 85.6 & 1.05 \\ \cline{1-1}\cline{3-5} 
LFC-MPC-FCBF &  & 20.8 & 236.5 & 0.48 \\ \hline
\end{tabular}
\label{tab1}
\end{table}

\begin{table}[htbp]
\caption{Statistical Results for Scenario 2}
\centering
\scriptsize
\setlength{\tabcolsep}{3pt} 
\renewcommand{\arraystretch}{1.2} 
\begin{tabular}{|c|c|c|c|}
\hline
\textbf{Metric} & \textbf{N} & \textbf{AAC-MPC-FCBF} & \textbf{LFC-MPC-FCBF} \\ \hline

\multirow{3}{*}{\textbf{Task Duration [s]}} 
& 5  & 6.73$\pm$0.12 & 7.41$\pm$0.21 \\ \cline{2-4}
& 10 & 6.72$\pm$0.05 & 7.32$\pm$0.15 \\ \cline{2-4}
& 15 & 6.68$\pm$0.02 & 7.02$\pm$0.06 \\ \hline

\multirow{3}{*}{\textbf{Computation Time [ms]}} 
& 5  & 39.4$\pm$2.36 & 183.3$\pm$5.64 \\ \cline{2-4}
& 10 & 136.3$\pm$6.28 & 516.7$\pm$15.65 \\ \cline{2-4}
& 15 & 540.8$\pm$12.55 & 4025$\pm$85.45 \\ \hline


\multirow{3}{*}{\textbf{Aggressive Driving Rate [\%]}} 
& 5  & 9.09 & 18.18 \\ \cline{2-4}
& 10 & 9.09 & 18.18 \\ \cline{2-4}
& 15 & 9.09 & 18.18 \\ \hline

\multirow{3}{*}{\textbf{Tracking Error [m]}} 
& 5  & 0.4899 & 0.4809 \\ \cline{2-4}
& 10 & 0.4765 & 0.4755 \\ \cline{2-4}
& 15 & 0.4586 & 0.4559 \\ \hline

\end{tabular}
\label{tab2}
\end{table}


figure/Fig.\ref{fig:AV results} and Table \ref{tab2} present the experimental results of cooperative lane-changing involving 15 vehicles. Compared to LFC, AAC consistently outperformed LFC in computation time across various MPC time horizons ($N=5, 10, 15$). AAC's computation times were $4.6$ to $7.4$ times faster than LFC's. Here, aggressive driving behavior is defined as the vehicle's sharp turns or sudden acceleration which typically increases the risk of collisions. As illustrated in figure/Fig. \ref{fig:AV results} (2) and (3), AAC method demonstrates smoother trajectories with fewer abrupt turns compared to LFC approach, indicating a lower aggressive driving rate. This enhanced smoothness can be attributed to the fundamental fairness principle of AAC, which provides each vehicle with a more equitable opportunity to plan its path.  Furthermore, figure/Fig. \ref{fig:AV results} (1) reveals that AAC maintains a larger minimum distance from the uncontrolled pedestrian (who holds the highest priority on the road), thereby significantly improving safety performance. 
While not superior in tracking precision, AAC excels in computational efficiency and safety, especially in complex, multi-robot environments.

\subsection{Safety Evaluation: CBF vs. F-CBF}
This section evaluates the safety performance of different control barrier functions in an emergency braking scenario. In the experiment, both the ego robot and the uncontrolled robot move in a straight line at same initial speed, with the uncontrolled robot positioned 7 meters ahead of the ego robot initially. When the uncontrolled robot executes an emergency stop, decelerating to 0 m/s, the obstacle avoidance responses of the ego robot using CBF and F-CBF strategies are compared to assess the safety advantages of F-CBF.

\begin{figure}
    \centering
    \includegraphics[width=3 in]{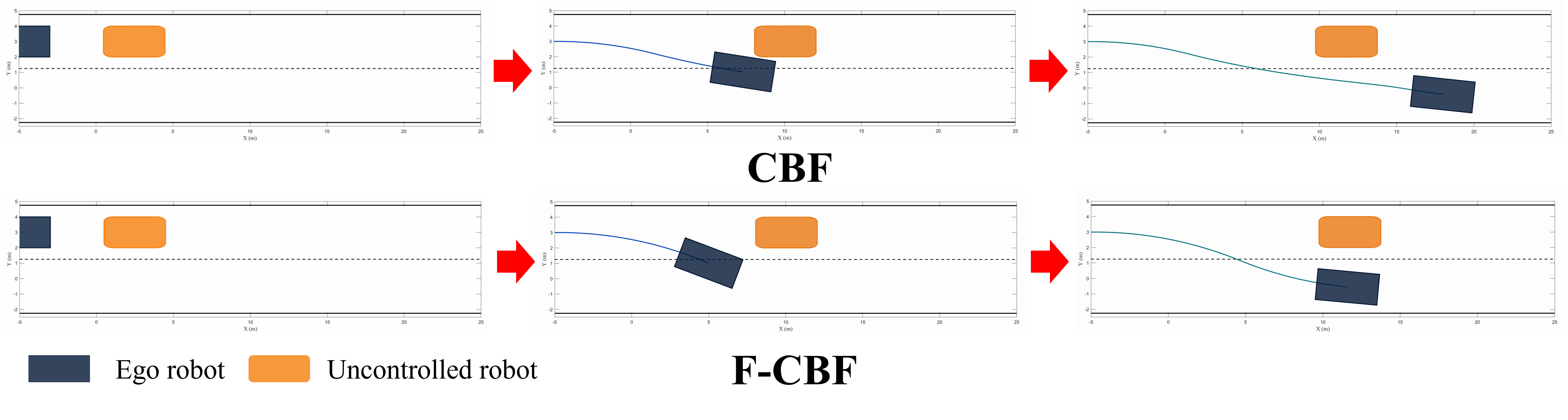}
    \caption{Three snapshots of the experiment of robustness analysis}
    \label{fig:CBF-FCBF}
\end{figure}

We evaluate using three metrics: infeasibility rate, minimum distance, and time-to-collision (TTC). The infeasibility rate shows how often the controller fails to find a solution, indicating the strategy's robustness. The minimum distance measures the closest Euclidean distance between two robots. TTC estimates time to potential collision based on the robot's speed and path. A higher average TTC indicates better performance, as it shows the strategy maintains a larger safety margin and reacts earlier to potential obstacles. Table \ref{tab3} shows how CBF and F-CBF perform on safety. First, in terms of the infeasibility rate, CBF often fails to find solutions when speeds are higher. In contrast, F-CBF, with its flexible boundary, ensures feasible solutions in a wider range of situations, demonstrating greater robustness. Additionally, F-CBF maintains a larger minimum distance (figure/Fig. \ref{fig:CBF-FCBF} shows F-CBF helps the robot slow down faster, avoiding obstacles more effectively). Finally, the higher TTC further validates the advantage of F-CBF in obstacle avoidance, highlighting its superior performance in terms of safety and response efficiency.

\begin{table}[htbp]
\caption{Comparisons between CBF and F-CBF}
\centering
\footnotesize
\setlength{\tabcolsep}{3pt}
\begin{tabular}{|c|c|p{1.5cm}|p{1.5cm}|p{1.5cm}|}
\hline
\textbf{Method} & \textbf{Init. Speed (m/s)} & \textbf{Infeasibility Rate} & \textbf{Min. Dist. [m]} & \textbf{Average TTC [s]} \\ \hline

\textbf{CBF} & \multirow{2}{*}{5}  & 43.3\% & 1.50 & 2.04 \\ \cline{1-1}\cline{3-5} 
\textbf{F-CBF} &  & 0.0\%  & 1.65 & 2.10 \\ \hline

\textbf{CBF} & \multirow{2}{*}{10} & 33.3\% & 0.87 & 2.13 \\ \cline{1-1}\cline{3-5} 
\textbf{F-CBF} &  & 0.0\%  & 1.05 & 3.55 \\ \hline

\textbf{CBF} & \multirow{2}{*}{15} & 100.0\% & \textemdash & \textemdash \\ \cline{1-1}\cline{3-5} 
\textbf{F-CBF} &  & 0.0\%  & 4.04 & 5.62 \\ \hline

\end{tabular}
\label{tab3}
\end{table}


\section{Conclusion and Future Work}

We propose an Alternating Authority Control (AAC) framework for multi-robot systems, which dynamically allocates control authority to prevent deadlocks and reduce inefficiencies. Additionally, we propose the Flexible Control Barrier Function (F-CBF) for handling dynamic obstacles and integrate the AAC framework with the MPC-FCBF strategy, creating a new safety-critical control system, AAC-MPC-FCBF. Simulations demonstrate that AAC-MPC-FCBF reduces computation time while improving robustness and fairness, especially in densely clustered robot formations. Future work will extend this framework to address model uncertainties and measurement noise. We also plan to validate the AAC-MPC-FCBF system on underwater vehicles and drones to enhance its real-world reliability.


\bibliographystyle{IEEEtran}
\bibliography{references}

\end{document}